\documentclass[journal]{IEEEtran}
\usepackage{cite}
\ifCLASSINFOpdf
   \usepackage[pdftex]{graphicx}\else\usepackage[dvips]{graphicx}\fi
\usepackage[cmex10]{amsmath}
\usepackage{amsfonts, amsthm, amssymb}
\usepackage[ruled]{algorithm}
\usepackage{algorithmic}
\usepackage{array}
\usepackage{mdwmath}
\usepackage{mdwtab}
\usepackage{eqparbox}
\usepackage{stfloats}
\usepackage{multirow}
\usepackage[caption=false,font=footnotesize]{subfig}
\usepackage{xcolor}
\usepackage{color,soul}
\usepackage{url}

\newtheorem{theorem}{Theorem}
\newtheorem{definition}{Definition}
\newtheorem{lemma}{Lemma}
\newtheorem{proposition}{Proposition}

\newcommand{\mf}{\mathbf}
\newcommand{\mb}{\mathbb}

\newcolumntype{M}[1]{>{\centering\arraybackslash}m{#1}}

\begin{document}
\title{Failure-averse Active Learning for Physics-constrained Systems}

\author{Cheolhei Lee,
        Xing Wang,
        Jianguo Wu,  and~Xiaowei~Yue,~\IEEEmembership{Senior Member,~IEEE}
        
\thanks{Manuscript received 00 00, 2021; revised 00 00, 2021. \textit{(Corresponding author: Xiaowei Yue.)}}

\thanks{C. Lee and X. Yue are supported by the US National Science Foundation under CMMI-2035038.}

\thanks{Mr. Cheolhei Lee is a Ph.D. student in the Grado Department of Industrial and Systems Engineering, Virginia Tech; Dr. Xiaowei Yue is an assistant professor in the Grado Department of Industrial and Systems Engineering, Virginia Tech, Blacksburg, VA, 24061 USA (e-mail: cheolheil@vt.edu; xwy@vt.edu).}

\thanks{Dr. Xing Wang is an assistant professor in the Department of Mathematics, Illinois State University, Normal, IL, 61790 USA.}
\thanks{Dr. Jianguo Wu is an assistant professor in the Department of Industrial Engineering and Management, Peking University, Beijing, 100080 China.}

}

\maketitle

\begin{abstract}
Active learning is a subfield of machine learning that is devised for design and modeling of systems with highly expensive sampling costs. Industrial and engineering systems are generally subject to physics constraints that may induce fatal failures when they are violated, while such constraints are frequently underestimated in active learning. In this paper, we develop a novel active learning method that avoids failures considering implicit physics constraints that govern the system. The proposed approach is driven by two tasks: the safe variance reduction explores the safe region to reduce the variance of the target model, and the safe region expansion aims to extend the explorable region exploiting the probabilistic model of constraints. The global acquisition function is devised to judiciously optimize acquisition functions of two tasks, and its theoretical properties are provided. The proposed method is applied to the composite fuselage assembly process with consideration of material failure using the Tsai-wu criterion, and it is able to achieve zero-failure without the knowledge of explicit failure regions.
\end{abstract}

\begin{IEEEkeywords}
Active Learning, Physics-integrated Machine Learning, Composite Structures Assembly
\end{IEEEkeywords}

\IEEEpeerreviewmaketitle
\section{Introduction}\label{se:Introduction}
\IEEEPARstart{A}{ctive} learning is a subfield of machine learning that maximizes information acquisition to reduce the labelling cost in supervised learning \cite{settles2009active}. Contrary to passive learning such as factorial, maximum entropy, Latin Hypercube design (LHD) \cite{santner2003design}, active learning optimizes the acquisition function that quantifies the potential importance of unlabelled data, and interactively queries the most informative design point to the oracle. Generally, acquisition functions refer to the up-to-date model or labelled data, while there are various strategies that can be adopted according to the preference in information criteria and characteristics of systems. The engineering domain is one of beneficiaries of active learning due to the high complexity of systems and the expensive cost of evaluation \cite{chen2020generative, yue2020active, zhang2019reif}.


However, active learning in engineering applications has been mostly utilized without considering coexisting or inherent constraints that may have different processes thereof. It is very crucial to consider such constraints in engineering systems, since most of them are subject to physics constraints that may induce fatal and irreversible failures. For example, design of the automatic shape control system in composite aircraft manufacturing need to consider potential material failures such as crack, buckling, and delamination caused by intolerable inputs \cite{wen2019virtual}. The inverse partial differential equation (PDE) problem is another example that is used to calibrate parameters in physics models based on observations. It usually involves physics constraints comprised of different PDEs, and the constraints must be satisfied in order to build first principled models. In both cases, the application of active learning without considering physics constraints may induce fatal failures, so constraints must be considered in order to avoid failures in the application of active learning in engineering problems.

To consider physics constraints in active learning, it is straightforward to define a safe region where design points satisfy safe conditions (with high probability at least), and conduct active learning within the safe region. However, it is not always possible in practice, since physics constraints cannot be explicitly attained due to the complex nature of the system. Examples include the crash damage analysis of commercial vehicles composed of different materials, and physicochemical interactions in corrosion of alloys. In these cases, fundamental physics laws and equations cannot be directly applied or are insufficient to accommodate the complexities of mechanisms. Physics-based numerical methods such as finite difference methods and finite element methods (FEM) \cite{moaveni2011finite} are well-established and convincing to analyze large classes of complex structures including failures, while they are too computationally demanding to identify the entire safe region. Moreover, their deterministic solutions are vulnerable to various uncertainty sources such as material properties, geometries, and loads.

In order to circumvent aforementioned limitations of physics-based approaches, machine learning models have been widely used in the engineering domain due to their flexibility, inexpensive prediction, and capability of uncertainty quantification (UQ). Physics information can be highly advantageous for machine learning in several aspects such as generalization and physical consistency \cite{willard2020integrating}. Especially, Gaussian processes (GPs) have shown remarkable performance in stochastic analysis of structural reliability that aims to evaluate the probability of system failure \cite{choi2006reliability}. A common reliability analysis approach employs the GP surrogate model of performance function associated with the system failure, and uses the acquisition function (e.g., \cite{bichon2008efficient, echard2011ak}) that leads to sampling near the boundary of safe and failure regions. The boundary is called the limit-state, which is the margin of acceptable structural design. However, the reliability analysis is mainly interested in the response surface associated with failure, and it is time-consuming due to the requirement of a large number of samples to estimate the underlying distribution at the limit-state. Hence, it can be data-inefficient to implement reliability analysis prior to the estimation of safe region. So the development of flexible active learning that takes account of target and failure processes simultaneously is promising for systems with implicit physics constraints.


The principle of active learning is exploitation of knowledge from observations, and exploration by tackling the knowledge such as the design point with maximum entropy or the most disagreeable point in the set of hypotheses. However, if unknown constraints exist in the design space, active learning can be very challenging, since the most informative design point may be located in the unknown failure region. Conversely, if active learning is too conservative to avoid failures, the resulted model will be vulnerable to unexplored safe regions. Consequently, active learning for physics-constrained systems should simultaneously take into account the following objectives: 
\begin{enumerate}
    \item maximizing the information acquisition for the target model;
    \item expanding explorable safe region by focusing on the constraint function,
\end{enumerate}
and they are must be achieved safely. Definitely, two objectives are at odds since they are associated with different functions, so the active learning strategy must be judiciously controlled. In this paper, we propose an active learning methodology for systems that are constrained by unknown failure processes. Our proposed method mainly focuses on the target process, while it simultaneously considers the failure process based on the GP model informed by physics-based models. The GP is adopted to incorporate uncertainty in the prediction of failure model and reduce the computational cost of the physics-based models. Two confidence levels are considered for safe exploration and safe region expansion, and a new acquisition function is developed to integrate two objectives of which the policy can be flexibly adjusted by the decision maker's preference. Our contributions in this paper are as follows.
\begin{enumerate}
    \item We develop the safe active learning strategy under the regime of implicit physics constraints in the system.
    \item A new acquisition function is proposed to enable flexible exploration and exploitation in the constrained design space by referring to two heterogeneous processes.
    \item Theoretical properties of the proposed method are provided to guarantee the performance.
\end{enumerate}


This paper is organized as follows. In section \ref{se:literature_review}, we review literature related to active learning applications in which physics involved and machine learning with unknown constraints. In section \ref{se:methodology}, we elucidate our active learning strategy considering implicit constraints to avoid failures. Section \ref{se:simulation_study} illustrates that how the proposed strategy works under the regime of unknown constraints with the simulation study. The real-world application to predictive modeling of composite fuselage deformation considering structural failure is presented in section \ref{se:case_study}. Lastly, a summary of this paper is provided in section \ref{se:conclusion}.


\section{Literature Review}\label{se:literature_review}
In this section, we discuss related literature dichotomizing into (i) active learning for engineering systems; and (ii) machine learning with unknown constraints. While the literature of two topics may not be related, the integration of two approaches can be the cornerstone of our approach. In the application of active learning to engineering systems, we focus on that how physics in the engineering system influences active learning. In machine learning with unknown constraints, we do not restrict the constraints therein being related to physics, yet focus on the ways of considering unknown constraints in machine learning. Since there are not many references whose main contribution lies in active learning, we extend our attention to more general machine learning approaches including the Bayesian optimization, and the inverse design in structural reliability analysis.

\subsection{Active Learning for Engineering Systems}
In the engineering domain, active learning is vastly utilized for surrogate modeling of expensive-to-evaluate systems, and the limit-state estimation in the structural reliability analysis. For the surrogate modeling, active learning aims to sample design points that minimizes the generalization error of the model. One of the most applied area is computational physics that often involves surrogate models of systems of which physics-based models are costly or absent. For response surface method (RSM), Alaeddini \textit{et al.} \cite{alaeddini2019sequential} proposed an active learning strategy adopting the variance reduction of Laplacian regularized parameters. Since RSM is often too restrictive to approximate complex pheomena, deep neural networks (DNNs) and GPs are substantially used for the surrogate modeling of PDE-based models. For DNNs, Costabal \textit{et al.} \cite{sahli2020physics} proposed a physics-informed neural network for cardiac activation mapping. The neural network model is guided by a loss function that involves the physics equation, one of the most widely used approaches to infuse physics into machine learning. Thereby the resulted model can be physically consistent. In their active learning, they chose the design point with the largest uncertainty quantified by the network model. Lye \textit{et al.} \cite{lye2021iterative} proposed active learning for surrogate modeling of PDE solutions that chooses the next query point minimizing the cost function with the sequentially updated DNN model. In order to ensure the feasibility, they confined the explorable settings with the feasible region known a priori. Pestourie \textit{et al.} \cite{pestourie2020active} employed the ensemble neural network model for photonic-device model, and used the quantified uncertainty of network model for active learning. However, their strategy does not consider physically infeasible settings, and they may induce failures if there exist failure conditions in the design space.

For GP models, Yang \textit{et al.} \cite{yang2018physics, yang2019physics} proposed a physics-informed GP for the stochastic PDE simulator. The GP model is informed by replicated observations of the PDE simulator, and the predictive variance is referred for active learning. However, their strategy is unconstrained, so the next query point may not satisfy the preliminary conditions. Chen \textit{et al.} \cite{chen2020apik} developed the GP that incorporates linear and nonlinear PDE information. They involve the active learning strategy to determine the PDE points for their GP model by employing the integrated mean-squared error (IMSE) criterion \cite{santner2003design}. However, the criterion only refers to physics information that does not constrain the learning of the target process, so their approach also may lead to infeasible design points in the constrained system. Yue \textit{et al.} \cite{yue2020active} proposed the variance-based and the Fisher information criteria for GP considering uncertainty, and applied to the modeling of composite fuselage deformation. However, they also did not involve physics constraints during the learning process.

The GP is also frequently employed in structural reliability analysis to model the system reliability, and acquisition functions thereof give more weights on sampling from the vicinity of the limit-state. Echard \textit{et al.} \cite{echard2011ak} proposed an acquisition function to estimate the limit-state with GP models, and suggested the framework that encompasses the limit-state estimation and the Monte-Carlo simulation for density estimation. Bichon \textit{et al.} \cite{bichon2008efficient} proposed an acquisition function called expected feasibility function that quantifies the closeness of design points to the limit-state considering quantified uncertainty of the GP. More explicitly, the vicinity of the limit-state was defined with the predictive uncertainty of GP for every candidate design point. Bect \textit{et al.} \cite{bect2012sequential} proposed the stepwise uncertainty reduction approach that employs the IMSE criterion associated with the estimated safe region. Wang \textit{et al.} \cite{wang2014maximum} proposed the maximum confidence enhancement method whose acquisition function aggregates the distance to the limit-state, input density, and the predictive uncertainty by multiplying them. The components are the same to \cite{bichon2008efficient}, while the formulation of acquisition function is different. Sadoughi \textit{et al.} \cite{sadoughi2018sequential} proposed the dynamically adjustable acquisition function using smooth weighting the closeness unlike to the aforementioned approaches. However, these surrogate-based structural reliability analysis and their variations are free from failures, since they observe systems using simulators. Therefore, their approaches cannot be used in the presence of constraints.

\subsection{Machine Learning with Unknown Constraints}
Constraints are critical in every optimization problem that must be satisfied for every solution, although they may be unknown a priori in practice. It is important to consider constraints in machine learning as well, since it is one way to incorporate background domain knowledge and thereby make the model more consistent with reality. There are mainly two distinct manners in considering unknown constraints in machine learning. One is that any evaluation at an infeasible design point is never allowed, and the other is that constraints can be disregarded in the trajectory of solution. We refer the first type as safe exploration since any infeasible design is directly related to the system failure which is critical for continuing the system analysis. The second type is usually addressed in Bayesian optimization, which is a sequential design strategy for the global optimization of black-box functions.

For safe exploration, Schreiter \textit{et al.} \cite{schreiter2015safe} proposed active learning for GPs whose motivation is closest to ours. They used the nuisance functions of GP classifiers for constraints, and used the lower confidence interval to ensure safety during the maximum entropy based active learning. However, their approach only focuses on the entropy of target function, so it could be problematic if the constraint models are lack of accuracy due to the target process-focused samples. Furthermore, the nuisance function may distort the numerical information from the constraint observations that are usually related to closeness to the boundary of the safe region. Sui \textit{et al.} \cite{sui2018stagewise} proposed a Bayesian optimization algorithm with unknown constraints. Their principle is to expand and conquer. That is, they expand the safe region in the first phase, and then the ultimate learning objective is achieved within the disclosed safe region. However, their safe expansion can be subject to slow convergence to the ground-truth unless the Lipschitz constants of unknown constraint functions are known beforehand. Considering the limitation of the safe expansion, Turchetta \textit{et al.} \cite{turchetta2020safe} suggested the safe exploration for interactive machine learning that can be adopted for active learning and Bayesian optimization. Their principle is that the safe region expansion is conducted when the next query position lies in the uncertain safe region. However, their method mainly considers dynamic systems that involve decisions and state transitions, so their approach is unsuitable to ours.

In Bayesian optimization, the GP is mostly considered for the approximation of black-box function, and one of the most widely used acquisition function is the expected improvement (EI) function. The function is straightforwardly formulated upon tractable Gaussian distribution of GP models: it chooses the point with the greatest expectation of improvement in the optimization solution considering uncertainty. Schonlau \textit{et al.} \cite{schonlau1998global} multiplied the probability of constraint satisfaction (also called expected feasibility) to the EI function. Gelbart \textit{et al.} \cite{gelbart2014bayesian} employs the latent GP models of unknown constraints that may not be observable simultaneously with the target process. They also used the EI function weighted by expected feasibility induced by the latent GP model. Gramacy \textit{et al.} \cite{gramacy2010optimization} assumed infeasible designs also can be informative to the solution of problem, so they proposed the integrated expected conditional improvement that is also weighted by expected feasibility. Later, the augmented Lagrangian approach coupled with the EI function was proposed in \cite{gramacy2016modeling}. The Lagrangian penalty is optimized in the outer loop of the algorithm so that the solution can be led to the feasible region. Picheny \cite{picheny2014stepwise} suggested to choose the design point that induces the largest expected volume reduction, quantified by integrating the production of expected feasibility and probability of the improvement function. Hern\'andez \textit{et al.} \cite{hernandez2015predictive} proposed the predictive entropy search with constraints that refers to the expected entropy reduction at the minimum associated with observations from the objective function and constraints. It automatically focuses one of objective and constraints by merging them into the integrated entropy. Basudhar \textit{et al.} \cite{basudhar2012constrained} used the probabilistic support vector classifier to discriminate the safe region. Then, the probability of constraint satisfaction is used to weight the EI function as well. However, the aforementioned approaches for unknown constraints do not enforce sampling from the safe region, so they are incapable of safe exploration.


\section{Methodology}\label{se:methodology}
\begin{figure*}[t!]
    \centering
    \includegraphics[width=0.9\textwidth]{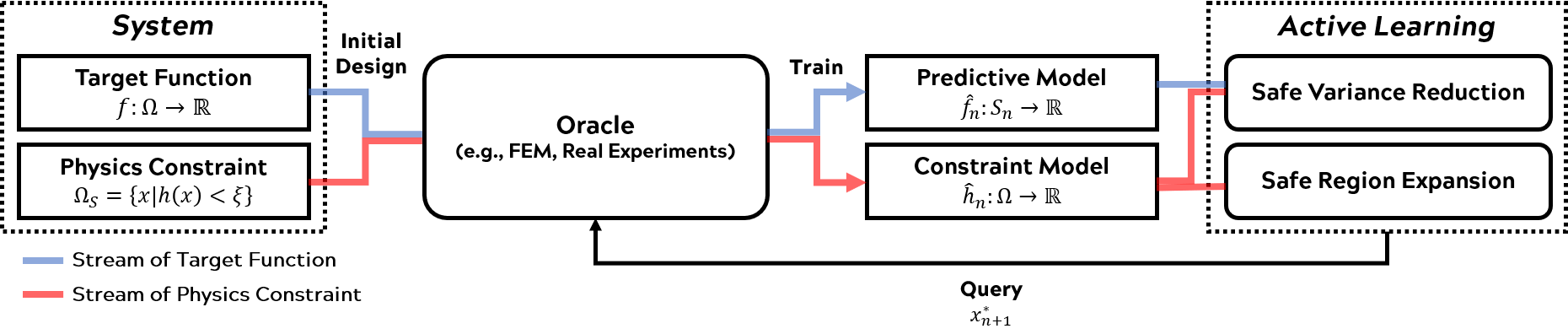}
    \caption{Overview of the proposed methodology}
    \label{fig:overview}
\end{figure*}

In this section, we propose our failure-averse active learning method for physics-constrained systems, the overview of which is shown in Fig. \ref{fig:overview}. The system of interest contains the target function, and there are governing constraints. The predictive modeling of target function is initialized with a passive design (e.g., space-filling), and the constrained model is introduced to inform the active learning module comprised of two objectives: the safe variance reduction and the safe region expansion. Then, both models are updated sequentially with data queried by the active learning strategy and labelled by the oracle. We begin with specifying our problem whose constraints can be evaluated along with the target process. Then, the safe exploration incorporating constraints is described in detail with its theoretical properties.

\subsection{Problem Statement and Gaussian Process Priors}
Consider a system defined over a compact and connected design space $\Omega \subseteq \mb{R}^D$. The system includes the target function $f:\Omega \to \mb{R}$, we want to predict, and the constraint function $h:\Omega\to\mb{R}$, related to the system failure and assumed to be independent of $f$. Let $\xi \in \mb{R}$ be a tolerable failure threshold associated with $h$, which should be defined conservatively considering intrinsic uncertainty of the process. For any design point $\mf{x}=\left[\, x_1\, \cdots\, x_D\, \right]^\top \in \Omega$, the system failure occurs when
\begin{align}
  h(\mf{x}) \ge \xi  \label{eq:const},
\end{align}
and safe otherwise. For example, $f$ can be the dimensional deformation of a solid structure given the force vector $\mf{x}$, and $h$ can be the resulted von Mises stress in the structure. Both functions can be observed or evaluated with the costly real experiment and relevant physics-based models as
\begin{align*}
    y = f(\mf{x}) + \epsilon_f,\qquad z = h(\mf{x}) + \epsilon_h,
\end{align*}
where $\epsilon_f\sim\mathcal{N}(0,\, \nu_f^2)$ and $\epsilon_h\sim\mathcal{N}(0,\, \nu_h^2)$ are observation noise. We assume that $\nu_f$ and $\nu_h$ are assumed to be known, while they can be also estimated with GP priors on $f$ and $h$ to be described later with nugget effects. The safe region $\Omega_{S}$, the subset of design space comprised of non-failure design settings, is unknown and difficult to obtain due to the prohibitive cost of the evaluation. We refer the complementary of safe region as the failure region such that $\Omega_{F} = \Omega\setminus \Omega_{S}$. 

In our case, due to the high cost of evaluation of $h$ and $f$, we prefer to sample the most informative set of design points that minimizes the generalization error associated with the target function. Suppose we have $n$ samples from the system, denoted by $\mathcal{D}_n = \{(\mf{x}_i,\,y_i,\,z_i)\}_{i=1}^n$, and $\hat{f}_n$ and $\hat{h}_n$ are our predictors of $f$ and $h$ trained with $\mathcal{D}_n$, respectively. Then, the expected risk of $\hat{f}_n$ using $L_p$ loss is
\begin{align*}
    \mathcal{R}_f(\hat{f}_n) = \int_{\Omega_{S}}L_p(f(\mf{x}),\; \hat{f}_n(\mf{x}))\;d\lambda (\mf{x}),
\end{align*}
where $L_p(y,\, y') = |y - y'|^p$ is the loss function, and $\lambda(\mf{x})$ is a probability measure defined over $\Omega$. In our problem, any violation of \eqref{eq:const} may incur prohibitive cost of failure in the system, so any $\mf{x}\in \Omega_{F}$ will not be considered for the system.

We assume that there exists a reproducing kernel Hilbert space (RKHS) for each of $f$ and $h$, and they are bounded therein. It allows us to model both functions with GPs with corresponding kernels such that $k_f:\Omega^2 \to \mb{R}$ and $k_h:\Omega^2 \to \mb{R}$ \cite{rasmussen2006gpml}. In this paper, we consider the automatic relevance determination using radial basis function (RBF) kernel for $\mf{x},\,\mf{x}'\in \Omega$ as
\begin{align*}
    k_f(\mf{x},\, \mf{x}') = \kappa_f^2 (\mf{x} - \mf{x}')^\top M^2_f(\mf{x} - \mf{x}') + \nu_f^2\delta(\mf{x},\,\mf{x}'),
\end{align*}
where $\kappa_f$ is the nonnegative scale hyperparameter, $M_f$ is the diagonal matrix of nonnegative length hyperparameters $\boldsymbol{\theta}_f=[\;\theta_{f,1},\;\ldots,\;\theta_{f,D}\;]^\top$, and $\delta$ is the Kronecker delta function for the nugget effect. By defining $k_h$ in the same manner, we can write $f$ and $h$ as
\begin{align*}
    f(\mf{x}) \sim \mathcal{GP}\left(\mu_f(\mf{x}),\; k_f(\mf{x},\, \mf{x}')\right),\\
    h(\mf{x}) \sim \mathcal{GP}\left(\mu_h(\mf{x}),\; k_h(\mf{x},\, \mf{x}')\right),
\end{align*}
where $\mu_f$ and $\mu_h$ are mean functions assumed to be zero without loss of generality.

Instead of employing a discriminative function for estimating the safe region, the GP regressor is more suitable for physics constraints since the output of $h$ is numerically informative. More explicitly, as $h(\mf{x})$ is closer to the failure threshold, we may notice that $\mf{x}$ is closer to the safe boundary. Moreover, discriminative functions require observations from both safe and failure regions, while regressors are not subject to such imbalance or absence of one class. Therefore, we fit our GP regression model directly on observed outputs from $h$, and refer to the distance between the output and the failure threshold to infer the probability of safety.

Let us denote $X_n$ as the $D\times n$ design matrix of $\left[\, \mf{x}_1\, \cdots\, \mf{x}_n\, \right]$, and $\mf{y}_n$ and $\mf{z}_n$ as the vector of $n$ observations from $f$ and $h$, respectively. With GP priors on $f$ and $h$, the hyperparameters $\Theta_f = \{\kappa_f,\,\boldsymbol{\theta}_f\}$ and $\Theta_h = \{\kappa_h,\,\boldsymbol{\theta}_h\}$ can be estimated by maximizing the log marginal likelihoods, which are
\begin{align*}
    \ell(\mf{y}_n|\,X_n,\Theta_f) &= -\frac{1}{2}\mf{y}^\top K_{f,n}^{-1} \mf{y} - \frac{1}{2}\log{|K_{f,n}|} - \frac{n}{2}\log 2\pi,\\
    \ell(\mf{z}_n|\,X_n,\Theta_h) &= -\frac{1}{2}\mf{z}^\top K_{h,n}^{-1} \mf{z} - \frac{1}{2}\log{|K_{h, n}|} - \frac{n}{2}\log 2\pi,
\end{align*}
where $K_{f,n}$ and $K_{h,n}$ are covariance matrices comprised of every pair of $\mf{x},\; \mf{x}'\in X_n$ given $\Theta_f$ and $\Theta_h$, respectively. Once $\hat{f}_n$ and $\hat{h}_n$ are obtained with maximizing their log marginal likelihoods, the predictive mean and variance of $\hat{f}_n$ at an unobserved design point $\mf{x} \in \Omega$ can be derived as
\begin{gather*}
    \mb{E}[\hat{f}_n(\mf{x})] = \mf{k}_f(\mf{x}, X_n) K_{f,n}^{-1} \mf{y}_n,\\
    \text{Var}\left(\hat{f}_n(\mf{x})\right) = k_f(\mf{x}) - \mf{k}_f(\mf{x}, X_n) K_{f, n}^{-1}\mf{k}_f(\mf{x}, X_n)^\top,
\end{gather*}
and so does $\hat{h}_n$'s.

\subsection{Safe Variance Reduction}
Let us consider $L_2$ loss called the mean squared error (MSE), while it is not required in practice for our approach. The IMSE criterion, we also adopt partially in our method, can be derived from the risk with $L_2$ loss \cite{cohn1996active}. Suppose $f_\ast$ is an unbiased predictor of $f$ with the minimum MSE (also called the best MSE predictor \cite{santner2003design}) with respect to $\Omega_S$ in the family of GP we consider. Then, the $L_2$ risk of $\hat{f}_n$ can be decomposed as
\begin{align}
\mathcal{R}_f(\hat{f}_n) &= \int_{\Omega_S}|f(\mf{x}) - \hat{f}_n(\mf{x})|^2\;d\lambda(\mf{x})\nonumber\\
&= \mathcal{R}_f(f_\ast) + \int_{\Omega_S}|f_\ast(\mf{x}) - \hat{f}_n(\mf{x})|^2\;d\lambda(\mf{x})\nonumber\\
& \ge \int_{\Omega_S}\text{Var}\left(\hat{f}_n(x)\right)d\lambda(\mf{x}),\label{eq:imse_predictor}
\end{align}
which is the sum of the $L_2$ risk of $f_\ast$ and the variance of $\hat{f}_n$. Since the $L_2$ risk of $f_\ast$ is negligible due to its unbiasedness, the generalization error with respect to the safe region can be reduced by focusing on the variance reduction in $\hat{f}_n$.

The IMSE criterion seeks the design point that is expected to reduce the variance of our predictor the most based on the realized $\hat{f}_n$. Let us denote the integrated variance of predictor in \eqref{eq:imse_predictor} as
\begin{align*}
    \mb{V}_{\Omega_S}(\hat{f}_n) = \int_{\Omega_S}\text{Var}\left(\hat{f}_n(\mf{s})\right) d\lambda(\mf{s}),
\end{align*}
where $\mf{s}\in \Omega_S$. Then, for an unobserved $\mf{x} \in \Omega$, the expected variance reduction over $\Omega_S$ in $\hat{f}_n$ given $\mf{x}$ is
\begin{gather}
    \Delta \mb{V}_{\Omega_S}(\hat{f}_{n}|\mf{x}) = \mb{V}_{\Omega_S}(\hat{f}_n) - \int_{\Omega_S}\text{Var}\left(\hat{f}_n(\mf{s}|\mf{x})\right) d\lambda(\mf{s}),\label{eq:imse}\\
    \text{Var}\left(\hat{f}_n(\mf{s}|\mf{x})\right)\!=\! k_f(\mf{s})\! -\! \mf{k}_f(\mf{s},\,X_{n+1})^\top K_{f,n+1}^{-1}\mf{k}_f(\mf{s},\,X_{n+1}),\label{eq:imse_integrand}
\end{gather}
where $X_{n+1}=[\,X_n\;\mf{x}\,]$, and $K_{f,n+1}$ is the covariance matrix of $X_{n+1}$. Eq. \eqref{eq:imse} is always nonnegative (Theorem 1 and 2 in \cite{chen2017sequential}) and $\mb{V}_{\Omega_S}(\hat{f}_n)$ is invariant to $\mf{x}$, so it suffices to maximize the second term. For the GP, choosing the next design point that minimizes \eqref{eq:imse} is called active learning Cohn (ALC), and it is widely used along with the active learning Mckay (ALM) that chooses the next query point with maximum entropy \cite{seo2000gaussian}. In terms of computational cost, the IMSE criterion is dominated by the inversion of $K_{f,n+1}$, so we adopt the Cholesky update for IMSE criterion in \cite{lee2021partitioned} to alleviate the cost and improve the numerical stability.

Unfortunately, \eqref{eq:imse} cannot be used directly, since the safe region is unknown a priori. Thus, we stick to our predictor $\hat{h}_n$ to estimate the safe region as
\begin{align}
    S_n = \{\mf{x}\in\Omega|\;\hat{\mu}_n^h(\mf{x}) + \beta_n\hat{\sigma}_n^h(\mf{x}) < \xi\},\label{eq:safe_region}
\end{align}
where $\hat{\mu}_n^h(\mf{x})$ and $\hat{\sigma}_n^h(\mf{x})$ are the mean and standard deviation of $\hat{h}_n(\mf{x})$, and $\Phi(\beta_n) = \text{Pr}(\hat{h}_n(\mf{x}) < \xi) = 1-p_n$ for which $p_n \in (0, 1)$. That is, $\beta_n$ is related to the probability of failure of $\mf{x}\in S_n$ as shown in the following theorem.
\begin{theorem}[Failure Probability]\label{theo:fail_prob}
Suppose we have $N$-sampling budget and $\hat{h}_n$. For $\gamma \in (0, 1)$, choosing design points $\mf{x}_i$'s from the safe region $S_i$ for $i\in\{n+1,\,\ldots,\,n+N\}$ has the failure probability as
\begin{align*}
    \text{Pr}\left(\inf_i \left(\hat{h}(\mf{x}_i) \ge \xi\,\big|\,\mf{x}_i \in S_{i-1}\right)\right) \le \gamma,
\end{align*}
where $S_i$ of which $\beta = \Phi^{-1}\left(1- \frac{\gamma}{N}\right)$.
\end{theorem}
The proof is provided in Appendix \ref{append:proof_theo1}. By constraining our choice of next design point $\mf{x} \in S_n$, \eqref{eq:imse} can be written as
\begin{align}
    \Delta\mb{V}_{\Omega_S}(\hat{f}_n|\mf{x})=&\mb{V}_{\Omega_S\setminus S_n}(\hat{f}_n) + \Delta\mb{V}_{S_n}(\hat{f}_n|\mf{x}),\label{eq:imse2}
\end{align}
where $\mf{s}\in S_n$. The first term of \eqref{eq:imse2} indicates the irreducible variance induced by discrepancy between $\Omega_S$ and $S_n$. Meanwhile, the second term is the reducible variance in the estimated safe region. Eq. \eqref{eq:imse2} shows that the choice of $p_n$ for $S_n$ affects the learnability of active learning and the safety. More explicitly, $S_n$ needs to be conservative to prevent failure by setting $p_n$ small enough, while too conservative setting of $S_n$ will increase the irreducible variance term in \eqref{eq:imse2} and reduce the explorable region. 

To extend the purview of variance reduction by $\mf{x}$ in \eqref{eq:imse2} while keeping safety, we may consider another safe region called the progressive safe region that has a more generous safety level than $S_n$ as
\begin{align*}
    S^+_n = \{\mf{x}\in\Omega|\;\hat{\mu}_n^h(\mf{x}) + \beta_n^+\hat{\sigma}_n^h(\mf{x}) < \xi\},
\end{align*}
where $\Phi(\beta_n^+) = \text{Pr}(\hat{h}_n(\mf{x}) < \xi) > 1 - p^+$ of which $p^+_n < p_n$. Straightforwardly, we have $S_n \subset S_n^+ \subseteq \Omega$. By considering $S^+_n$ as the reference set for the integrand \eqref{eq:imse_integrand}, i.e., $\mf{s}\in S^+_n$, we can reduce the irreducible variance in the first term. Consequently, we have the following acquisition function
\begin{align}
    J_f(\mf{x}) = \int_{S_n^+}\text{Var}\left(\hat{f}_n(\mf{s}|\mf{x})\right)\; d\lambda(\mf{s}),\label{eq:final_imse}
\end{align}
where $\mf{x}\in S_n$. While the increment of progressive safe region will reduce the unconsidered safe region $\Omega_S\setminus S_n^+$, it is noteworthy that it can also simultaneously increase the reducible variance in $\Omega_F\setminus S_n^+$, which is meaningless. Therefore, $S_n^+$ also need to be defined not too generous.

Even though $J_f$ is designed carefully with appropriate safe regions, we can minimize the irreducible variance by minimizing the discrepancy between $\hat{h}_n$ and $\Omega_S$. Generally speaking, $J_f$ only focuses on reducing the variance of $\hat{f}_n$, and does not care about reducing the discrepancy. In order to efficiently expand the explorable region and improve the estimation accuracy of safe region, we need to incorporate the information from $h$ as well as $f$ in the information criterion. In the following section, we illustrate the safe region expansion that focuses on the estimation of safe region boundary.

\subsection{Safe Region Expansion}
The safe region expansion is required to reduce the error induced by the mismatch of $S_n$ and $\Omega_S$, and furnish higher confidence in exploration. It is metaphorically similar to that we can win when we know much more about the opposite. To expand the safe region without failure, we need to exploit the numerically informative output of $h$ to approach to the boundary of safe region from inside thereof, and the expansion can be maximized when the design point is closest to the boundary \cite{castro2005faster}. Inspired by \cite{ranjan2008sequential}, we incorporate the uncertainty of $\hat{h}_n$ and closeness to the boundary with the following criterion:
\begin{gather*}
    I(\mf{x}) = \left\{\begin{array}{c c}\eta_n(\mf{x})^2 - (\hat{h}_n(\mf{x}) - \xi)^2 & \hat{h}_n(\mf{x}) \in (\xi - \eta_n(\mf{x}), \,\xi)\\0 & \text{Otherwise}\end{array}\right.,
\end{gather*}
where $\eta_n(\mf{x}) = \alpha \hat{\sigma}^h_n(\mf{x})$ of which $\alpha > 0$. $I(\mf{x})$ attains its maximum when $\hat{h}_n(\mf{x})=\xi$, which is the case of $\mf{x} \in \partial \Omega_S$. Otherwise, it gets additional scores when $\hat{h}_n(\mf{x})$ does not exceed the threshold within an acceptable interval. The role of $\alpha$ is magnifying the effect of uncertainty in $I(\mf{x})$. Let the expected value of $I(\mf{x})$ with respect to $\hat{h}_n$ be $J_h(\mf{x})$, which is the acquisition function for the safe region expansion expressed as
\begin{align}
    J_h(\mf{x}) =&\; \eta_n(\mf{x})^2\left(\hat{\Phi}^h_n(\xi) - \hat{\Phi}^h_n(\xi-\eta_n(\mf{x}))\right)\\& - \int_{\xi-\eta_n(\mf{x})}^{\xi}(h - \xi)^2\hat{\phi}^h_n(h)dh,\label{eq:safe_exp}
\end{align}
where $\hat{\Phi}^h_n$ and $\hat{\phi}^h_n$ are the CDF and the PDF of $\hat{h}_n(\mf{x})$. Eq. \eqref{eq:safe_exp} is composed of two terms: the first term is related to the uncertainty, and the second term is related to the closeness to the boundary. Consequently, maximizing \eqref{eq:safe_exp} leads to sampling near the boundary with high uncertainty if such points exist, and the most uncertain point otherwise. 

Obviously, interests of $J_f$ and $J_h$ are inherently different, since they are associated with different mechanisms, $f$ and $h$. Also, they are formulated for different purposes. It implies that the safe approximation of target function and the safe region expansion have trade-off, thus we need to compromise both criteria to determine the most informative design point. We discuss the framework for addressing the balancing between two criteria in the next section. 

\subsection{Harmonizing Acquisition Functions}
In this section, we integrate two acquisition functions to optimize (maximize) them judiciously to achieve the safe active learning. Two distinct acquisition functions are proposed to accomplish different objectives, and the optimization of two criteria is a multi-objective optimization (MOO) problem. Conceptually, we may think of a point $\mf{x}\in S_n$ that achieves the maximum of each criterion simultaneously, called as the utopia point. However, MOO typically has no single optimal solution contrary to the usual single-objective optimization. Therefore, the Pareto optimality concept is mostly referred to define the optimality in this regime. Let $\mf{J}(\mf{x}) = \left[\;J_f(\mf{x})\;J_h(\mf{x})\;\right]^\top$, then Pareto optimality and its weaker version are defined as follows.
\begin{definition}[Pareto Optimal] A point, $\mf{x}_\ast\in\Omega$, is Pareto optimal if and only if there does not exist another point, $\mf{x}\in \Omega$, such that $\mf{J}(\mf{x}) \le \mf{J}(\mf{x}_\ast)$, and $J_i(\mf{x}) < J_i(\mf{x})$ for at least one of $i\in\{f,\;h\}$.
\end{definition}
\begin{definition}[Weakly Pareto Optimal] A point, $\mf{x}_\ast\in\Omega$, is weakly Pareto optimal if and only if there does not exist another point, $\mf{x}\in \Omega$, such that $\mf{J}(\mf{x}) < \mf{J}(\mf{x}_\ast)$.
\end{definition}
Obviously, every Pareto optimal point is weakly Pareto optimal, while the reverse is not true. 

It is common to scalarize the vector-valued objective functions in MOO, and the formulation of problem is critical for Pareto optimality of the solution. In this paper, we use the weighted sum, which is widely used, to scalarize our criteria with the integrated acquisition function:
\begin{align}
    J(\mf{x}) = \left((1-w)J_f(\mf{x})^p + w J_h(\mf{x})^p\right)^{1/p},\quad w\in[0, 1],\label{eq:int_acqfun}
\end{align}
where $p\in \mb{N}$. The weight parameter $w$ in \eqref{eq:int_acqfun} exactly conveys the preference of decision maker between two objectives. For example, if one is more interested in the safe region expansion, the decision maker will weigh more on $J_h$, and decrease $w$ when the estimated safe region seems acceptable. Otherwise, some may begin with small $w$ to see if the safe region expansion is necessary. It turns out that the integrated acquisition function guarantees the Pareto optimality of its solution given $w$ as shown in the following proposition.
\begin{proposition}[Sufficient Pareto Optimality, \cite{marler2004survey}] For any $w\in(0, 1)$, a solution that maximizes the integrated criterion is Pareto optimal associated with $w$. When $w=0$ or $w=1$, a solution of the integrated criterion is weakly Pareto optimal associated with $w$.
\end{proposition}

However, $J_f$ and $J_h$ may be different in their scales, so the weight parameter $w$ cannot be determined straightforwardly. It is common to transform component objective functions in the formulation of MOO, so we may normalize each criterion as
\begin{align*}
    \overline{J_i}(\mf{x}) = \frac{J_i(\mf{x}) - \min{J_i}}{\max{J_i} - \min{J_i}},
\end{align*}
where minimum and maximum of $J_i$ for $i \in \{f,\, h\}$ stand for the minimum and maximum over $S_n$. To normalize objective functions, we need their maxima and minima, so we provide two scaling options in this paper. The first scaling method is for global searching in a dense-grid over the design space. By discretizing the design space into a dense-grid, we may evaluate all criteria over the grid. It can provide the heuristically global optimal solution, and make scaling more consistent. Another scaling method is to incorporate the lower and upper bounds of criteria. We already have that both criteria are nonnegative, so they are lower bounded by zero. For the upper bounds, let us consider the progressive safe region $J_f$ which is upper bounded by
\begin{align*}
    \mb{V}_{S_n^+}(\hat{f}_n) = \int_{S_n^+}\text{Var}\left(\hat{f}_n(\mf{s})\right)d\lambda(\mf{s}),
\end{align*}
since the expected variance reduction in \eqref{eq:final_imse} is nonnegative. Meanwhile, $J_h$ is upper bounded by
\begin{align*}
    \sup_{\mf{x}\in S_n}\eta_n(\mf{x})^2 = \sup_{\mf{x}\in S_n}\alpha^2\hat{\sigma}^h(\mf{x})^2
\end{align*}
from its original formulation. In this way, we can scale both criteria by their tractable bounds that can be obtained prior to the evaluation of every candidate.

The upper bounds of $J_f$ and $J_h$ can be referred to the asymptotic convergence of the integrated acquisition function in \eqref{eq:int_acqfun} as described in Theorem \ref{theo:asymptotic}, which is an important property of the proposed method.
\begin{theorem}[Asymptotic Convergence]\label{theo:asymptotic}
The integrated criterion converges to zero as the number of samples $n \to \infty$, which means the variance of target predictor is minimized, and the safe region is fully expanded.
\end{theorem}
The proof is given in Appendix \ref{append:proof_theo2}. Theorem \ref{theo:asymptotic} implies that the integrated criterion $J$ leads to the best estimator of $f$ and promising conservative estimation of $h$. Let us refer to our active learning as PhysCAL (\textbf{Phys}ics-\textbf{C}onstrained \textbf{A}ctive \textbf{L}earning), and its pseudocode is provided in Algorithm \ref{algo:physcal}. Note that PhysCAL can be terminated by not only the sampling budget, but also the prediction accuracy of the target model when the sampling budget is implicit or early stopping is reasonable. In order to do so, a separated testing dataset or cross-validation is required.

\begin{algorithm}[t!]
\caption{Active Learning for Physics-constrained Systems}
\begin{algorithmic}[1]\label{algo:physcal}
\STATE \textbf{Prerequisite}: $N\text{(Sampling budget)},\;\mathcal{D},\;\beta,\;\beta^+,\alpha,\;\;w$
\STATE Train $\hat{f},\,\hat{h}$ with $\mathcal{D}$
\WHILE{$N > 0$}
\STATE Evaluate $S_n,\;S_n^+$ over $\Omega$
\STATE $\mf{x}_\ast = \arg\max_{\mf{x}\in S_n} \overline{J}(\mf{x})$
\STATE Observe $y_\ast,\,z_\ast$ at $\mf{x}_\ast$
\STATE $N = N-1$
\STATE $\mathcal{D} = \mathcal{D}\cup \{\mf{x}_\ast,\,y_\ast,\,z_\ast\}$
\STATE Update $\hat{f},\,\hat{h}$ with $\mathcal{D}$
\STATE Update $\beta,\,\beta^+\,\alpha, w$ (Optional)
\ENDWHILE
\end{algorithmic}
\end{algorithm}


\section{Simulation Study}\label{se:simulation_study}
\begin{figure*}[t!]
\centering
\subfloat[]{\includegraphics[width=0.277\textwidth]{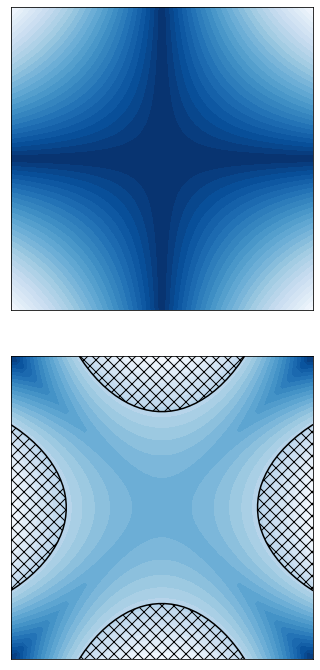}\label{fig:sim_gt}}\hfil
\subfloat[]{\includegraphics[width=0.28\textwidth]{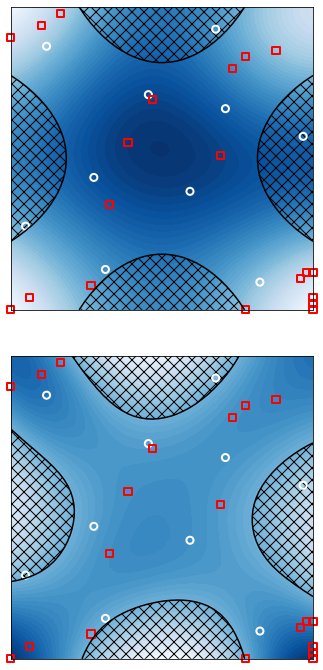}\label{fig:sim_physcal}}\hfil
\subfloat[]{\includegraphics[width=0.28\textwidth]{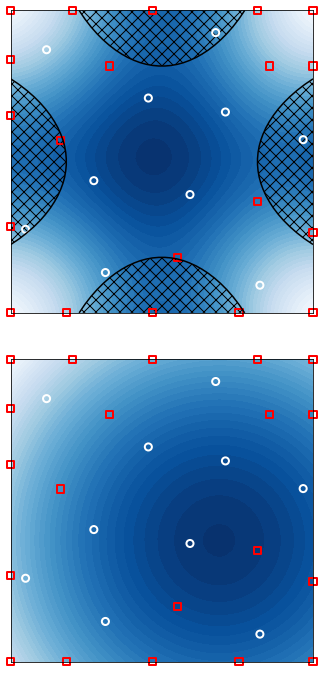}\label{fig:sim_schre}}
\caption{Simulation Study Result (a) Ground truths of target (top) and constraint (bottom) functions. (b) PhysCAL's (proposed method) estimation of target (top) and constraint (bottom). (c) Benchmark method's (Schreiter \textit{et al.} \cite{schreiter2015safe}) estimation of target (top) and constraint (bottom). Hatched regions in top figures are the true failure region, and the bottom of (b) is the estimated failure region. Note that the bottom of (c) has no estimated failure region. White circles are initial design points, and red squares are sampled with active learning.}
\label{fig:2d_sim}
\end{figure*}

In this section, we apply our active learning to the approximation of a constrained 2-D simulation function. The response surface of the target function is shown in the left top of Fig. \ref{fig:2d_sim}, and the constraint function's response surface and the failure region are in the left bottom thereof. We set the safe region as $\Omega_S = \{x\in[0,\;1]^2| h(x) < 0.7\}$, thereby the failure region ratio to the design space being approximately 0.28. Assuming we have no prior knowledge of the safe design settings, 10 initial samples are obtained over the design space using the maximin LHD, which yields 2-3 samples from the failure region in 10 replications. Observations from $f$ and $h$ are corrupted by Gaussian noise, and additional 20 samples are obtained with active learning. According to Theorem \ref{theo:fail_prob} and the sampling budget, we set $p = 0.001/20$ to retain the safety, and $p^+ = 0.01$ without update. As the benchmark method, the safe exploration for GP in \cite{schreiter2015safe} (referred as SEGP) is considered. For the other parameter settings, $\alpha=2$ and $w=\{0.0,\, 0.1,\, \ldots,\, 0.9,\,1.0\}$ are used.

In the simulation study, setting the safety level in the benchmark method as high as ours is impossible due to the insufficient number of samples to make a nontrivial explorable region as they mentioned in their work. Moreover, the benchmark method fails to estimate a plausible failure region, thereby resulting frequent failures as shown in the right of Fig. \ref{fig:2d_sim}. Meanwhile, our method estimated the failure region much better for every instance, and explored more safely as shown in the center of Fig. \ref{fig:2d_sim}.

\begin{figure}[!b]
    \centering
    \includegraphics[width=0.45\textwidth]{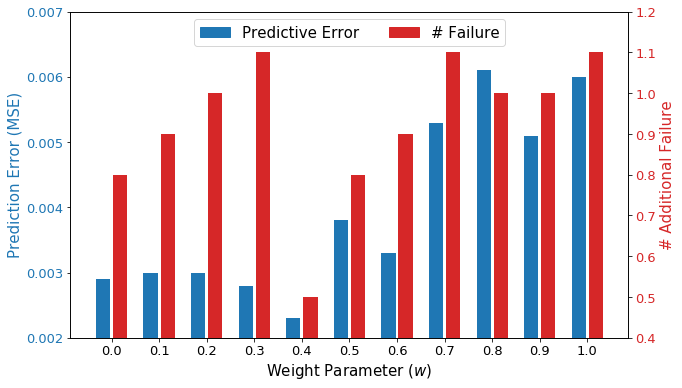}
    \caption{Performance of PhysCAL associated with different weight parameters in the simulation study}
    \label{fig:sim_w_sense}
\end{figure}

As a result, PhysCAL achieved the prediction error (MSE) 0.0023 with 0.5 additional number of failures on average, and there are five zero-failure cases of ten replications. Meanwhile, the benchmark method achieved a better prediction accuracy with an MSE of 0.001, but the averaged additional failures was 6.7 among 20 queries with no case of zero-failure. It is not surprising that the benchmark method did better in predictive accuracy, since exploration was unrestricted by underestimating the failure region. Hence, in the case that one failure is very crucial, our approach will be more suitable.

Fig. \ref{fig:sim_w_sense} shows the performance of PhysCAL with different weight parameter settings. We can observe that a low weight parameter may improve the prediction accuracy, but does not necessarily make our approach more conservative due to lack of knowledge in failure region. Increasing $w$ may compel our exploration being more aggressive with an increased number of failures, and induce low predictive accuracy due to indifference to the variance reduction in the target approximation. In this simulation, $w=0.4$ was the best choice among considered values with promising predictive accuracy and the least number of failures.

\section{Case Study}\label{se:case_study}
In this section, the proposed method is applied to the predictive modeling of deformation of the composite fuselage assembly in aerospace manufacturing. Composite materials such as carbon fiber reinforced polymers are extensively applied to various domains (including aerospace, automotive, construction and energy) due to its versatility, high strength-to-weight ratio, and corrosion-resistance. However, composite materials are nonlinear and anisotropic due to their structural natures \cite{hyer2009stress}, so flexible models and adaptive design of experiments are required to predict its deformation precisely. Moreover, since they are also subject to structural failures, manufacturers should avoid unsafe load settings in the fabrication of composite structures.

\begin{figure*}[!t]
    \centering
    \includegraphics[width=0.9\textwidth]{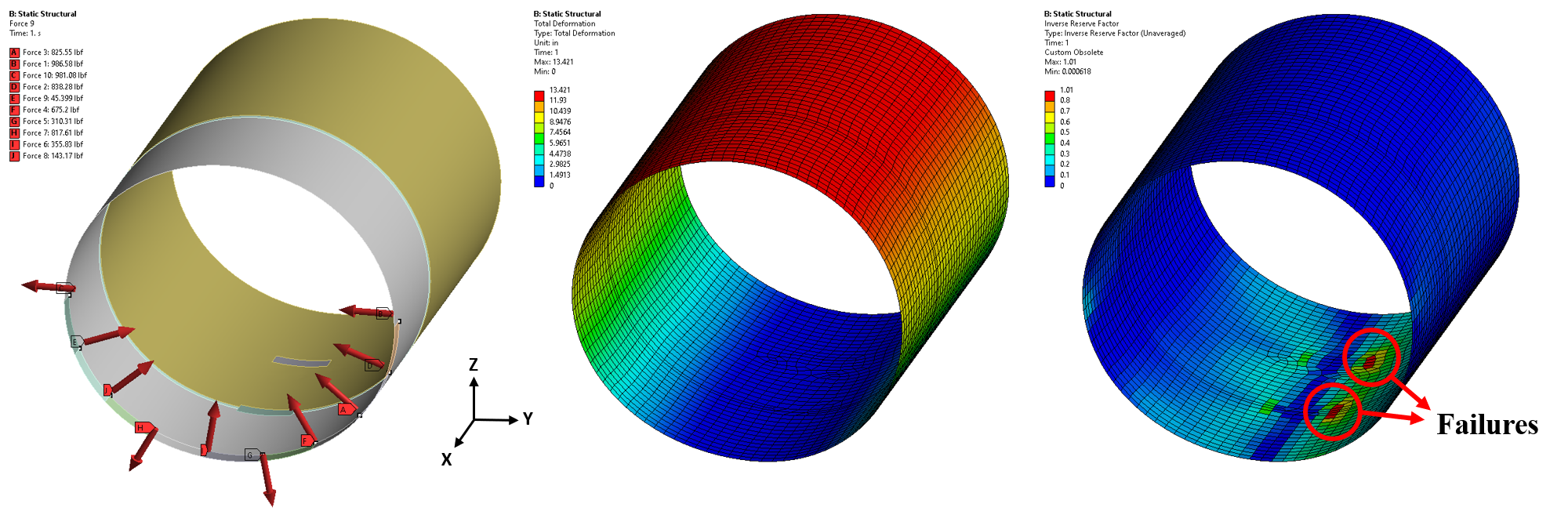}
    \caption{Shape control of composite fuselage in the FEM. (left) Actuator inputs. (center) Resulted deformation. (right) Resulted Tsai-wu criterion.}
    \label{fig:tsai_wu_fem}
\end{figure*}

In aircraft manufacturing, each fuselage section is subject to dimensional deviations at the joint rim due to the multi-batch manufacturing system. Consequently, to assemble fuselage sections, the shape control procedure is required to reshape them homogeneously. In the shape control procedure, the fuselage section is placed on the supporting fixture, and ten equispaced hydraulic actuators are introduced to reshape the fuselage as shown in the left figure of Fig. \ref{fig:tsai_wu_fem}. A promising approach for the optimal shape control procedure is to employ a highly precise predictive model of composite fuselage's deformation instead of time-consuming physics-based model \cite{yue2018surrogate, lee2020neural}. However, the construction of such model validated with real experiment is very challenging due to the expensive cost of sampling, and the risk of structural failure due to unsafe loads.

There are different types of failures in composite materials such as fiber, matrix, and ply failures. Likewise, a number of composite material failure criteria (e.g., Tsai-wu, Tsai-hill, Hoffman, Hashin) are devised for different modes of failures \cite{orifici2008review}. In this paper, we considered the Tsai-wu criterion, which is one of the most widely used interactive failure criterion. Note that it is possible to consider more than one criterion simultaneously by taking the most parsimonious criterion, or considering the intersection of safe regions defined by multiple constraint GP models. Briefly, Tsai-wu criterion considers interactions between different stress components in addition to the principal stresses (in a homogeneous element). Using the principal material coordinate system on the cubic element of composite material, consider three directions: 1 is the fiber direction; and 2 and 3 are  directions perpendicular to 1, respectively. Let $\sigma_{i}^T$ and $\sigma_{i}^C$ be the tensile failure stress and the compressive failure stress in $i\in\{1,\,2,\,3\}$ direction, and $\tau_{12}^F$ be the shear failure stress in the 12 plain. The Tsai-wu criterion is defined as,
\begin{align*}
    \left(\frac{1}{\sigma^T_1} - \frac{1}{\sigma^C_1}\right)\sigma_1 + \left(\frac{1}{\sigma^T_2} - \frac{1}{\sigma^C_2}\right)\sigma_2 + \frac{\sigma_1^2}{\sigma^T_1 \sigma^C_1} + \frac{\sigma_2^2}{\sigma^T_2\sigma^C_2} \\+\left(\frac{\tau_{12}}{\tau^F_{12}}\right)^2 - \frac{\sigma_1\sigma_2}{\sigma^T_1\sigma^C_1\sigma^T_2\sigma^C_2}\ge1,
\end{align*}
where the left-hand side is the nonnegative criterion value, and the failure occurs when it exceeds one. 

The Tsai-wu criterion value induced by the shape adjustment solved by the FEM is shown in the right of Fig. \ref{fig:tsai_wu_fem}. We can observe that failures are occurred at the bottom of the fuselage, since fixtures that sustain the fuselage are restricting its deformation. Not only limited to the Tsai-wu criterion, physics constraints have many assumptions such as homogeneity, absence of higher-order interactions, etc., although they are convincing apparatuses to consider the structural reliability. Hence, they are typically utilized with the safety-of-margin (the reciprocal of acceptable failure criterion) or UQ to prevent unexpected failures. Likewise, the safe shape control system should consider the failure criterion not only its value, but also the additional safety measures.

\subsection{Experiment Settings}
The proposed methodology is applied to the construction of predictive model of deformation. A well-calibrated FEM simulator of the procedure is referred to as our oracle considering the risk of real experiment. The target function's input is the vector of unidirectional forces (in lbf) of ten actuators, and the output is ${Y,\,Z}$-directional deformations (in microinch) of fuselage at one of 91 critical points around the rim. The maximum magnitude of actuator force is 1,000 lbs, which may cause failures in the structure (see Fig. \ref{fig:tsai_wu_fem}). As our additional safety measure, the margin of safety with Tsai-wu criterion is set with 1.25 (i.e., the acceptable criterion is 0.8).

For the initial design, the maximin LHD is used to generate 20 observations for both deformation and failure criterion, and additional 20 samples are queried by different methods: random, ALM, ALC, SEGP, and PhysCAL. We adopted the pool-based scenario in this case study by providing the 400 size of candidate pool that uniformly spreads out the design space. Considering the variability in the initial design, we have generated ten initial dataset independently, and replicated the experiment.

For PhysCAL, we also considered different weight parameters as the simulation study, and set other parameters as $\alpha=2$, $p=0.001/20$, and $p^+=0.1$. In SEGP, we reduced the safety level of which from the PhysCAL's until that SEGP induces a nonempty explorable space. For the model evaluation, we used 100 safe samples as the testing dataset that is independently generated with the candidate pool, and the mean absolute error (MAE) is used as the metric.

\subsection{Result}
The result is summarized in Table \ref{tab:case_result}. First, we can observe that PhysCAL outperforms other methods in the number of additional failures. It achieves nine zero-failures from ten cases. Also, we can observe that ALM, ALC, and SEGP incurred more failures than the random. The reason is that the design space is almost dominated by the safe region, but the failure region may be more interesting than elsewhere. Interestingly, SEGP is inapplicable in this case when the initial dataset does not contain failure samples, since the method uses the binary classifier. It implies that employing the regressor as the constraint model is more advantageous when the prior information has no failure.

In terms of prediction accuracy, PhysCAL performs much better than random sampling, and comparable to other active learning approaches considering the scale of metric. We can conjecture that other methods are able to observe from the failure region that may be informative, so their accuracy is the consequence of unsafe exploration. Furthermore, PhysCAL is more flexible than other methods, since we may update the weight parameter in PhysCAL during data acquisition to focus more on the variance reduction as well as other methods. Therefore, PhysCAL is more promising for this case considering the risk of failure in the system.
\begin{table*}[!t]
\caption{Result of case study}
\label{tab:case_result}
\centering
\begin{tabular}{M{0.045\textwidth}|M{0.058\textwidth}M{0.08\textwidth}|M{0.058\textwidth}M{0.08\textwidth}M{0.058\textwidth}M{0.08\textwidth}|M{0.058\textwidth}M{0.08\textwidth}|M{0.058\textwidth}M{0.08\textwidth}}
\hline\hline
\multirow{2}{*}{\textbf{CASE}} & \multicolumn{2}{c|}{\textbf{Random}} & \multicolumn{2}{c}{\textbf{ALM}} & \multicolumn{2}{c|}{\textbf{ALC}} & \multicolumn{2}{c|}{\textbf{SEGP} \cite{schreiter2015safe}} & \multicolumn{2}{c}{\textbf{PhysCAL}} \\
                      & MAE       & \# Add. Fail    & MAE     & \# Add. Fail  & MAE     & \# Add. Fail   & MAE        & \# Add. Fail      & MAE       & \# Add. Fail    \\ \hline
1                     & 6.303     & 1               & 2.573   & 4             & 2.572   & 4              & \multicolumn{2}{c|}{N/A}       & 2.640     & 0               \\
2                     & 10.135    & 0               & 2.572   & 4             & 2.572   & 3              & \multicolumn{2}{c|}{N/A}       & 2.639     & 0               \\
3                     & 7.975     & 1               & 1.628   & 5             & 1.635   & 4              & \multicolumn{2}{c|}{N/A}       & 3.530     & 0               \\
4                     & 2.368     & 0               & 2.862   & 4             & 2.143   & 4              & \multicolumn{2}{c|}{N/A}       & 3.667     & 0               \\
5                     & 2.427     & 1               & 2.161   & 4             & 1.939   & 5              & \multicolumn{2}{c|}{N/A}       & 2.465     & 0               \\
6                     & 4.570     & 0               & 2.572   & 4             & 2.572   & 3              & \multicolumn{2}{c|}{N/A}       & 2.639     & 0               \\
7                     & 6.163     & 0               & 2.344   & 4             & 2.631   & 4              & \multicolumn{2}{c|}{N/A}       & 2.387     & 0               \\ \hline
8                     & 6.704     & 1               & 1.427   & 4             & 1.272   & 4              & 1.892      & 0                 & 3.487     & 0               \\
9                     & 3.326     & 0               & 2.009   & 4             & 1.900   & 3              & 2.631      & 1                 & 2.286     & 0               \\
10                    & 4.602     & 0               & 1.970   & 4             & 1.583   & 4              & 2.369      & 2                 & 2.585     & 1               \\ \hline
\begin{tabular}[c]{@{}c@{}}Mean\\ (Std.)\end{tabular} &
  \begin{tabular}[c]{@{}c@{}}5.383\\ (2.345)\end{tabular} &
  \begin{tabular}[c]{@{}c@{}}0.4\\ (0.5)\end{tabular} &
  \begin{tabular}[c]{@{}c@{}}2.244\\ (0.470)\end{tabular} &
  \begin{tabular}[c]{@{}c@{}}4.1\\ (0.3)\end{tabular} &
  \begin{tabular}[c]{@{}c@{}}2.046\\ (0.441)\end{tabular} &
  \begin{tabular}[c]{@{}c@{}}3.8\\ (0.6)\end{tabular} &
  \begin{tabular}[c]{@{}c@{}}2.297\\ (0.305)\end{tabular} &
  \begin{tabular}[c]{@{}c@{}}1.0\\ (0.8)\end{tabular} &
  \begin{tabular}[c]{@{}c@{}}2.832\\ (0.518)\end{tabular} &
  \begin{tabular}[c]{@{}c@{}}0.1\\ (0.3)\end{tabular} \\ \hline\hline
\end{tabular}
\end{table*}

\subsubsection{Weight Parameter} 
\begin{figure}[!t]
    \centering
    \includegraphics[width=0.45\textwidth]{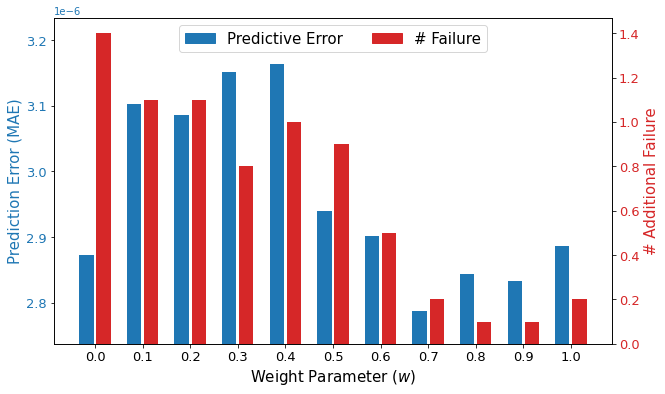}
    \caption{Performance of PhysCAL associated with different weight parameters in the case study}
    \label{fig:w_sense}
\end{figure}

Different weight parameters are considered for PhysCAL in the case study, and the performance of different weight parameter settings is provided in Fig. \ref{fig:w_sense}. In terms of safety, increasing weight parameters can improve the safety of active learning in this case. That is, even though we set $S_n$ extremely safe as in this case study, the safe region expansion is critical to the safe exploration. However, we also observe that setting $w=1.0$ also increases the number of failure slightly, so we should not make PhysCAL too aggressive by only focusing on the safe region expansion. For the predictive accuracy, we also observe that any extreme setting is not an optimal solution. 

There are more related factors that may affect the significance of weight parameter. The failure region in this case is much smaller than the safe region, thus quite aggressive exploration may be acceptable (i.e., increasing $w$). In the simulation case, the failure region ratio is higher than this case, so $w=0.4$, which is smaller than the best weight value of the case study, perform better. Also, the correlation between target and constraint functions can be a reason. More explicitly, we know that the safe region expansion term includes the predictive uncertainty term. Thus, when $f$ and $h$ are correlated, the safe region expansion term also can be helpful to improve the predictive accuracy of target model. In this case, increasing $w$ is relevant.

\subsection{Margin of Safety} In order to observe the effect of margin of safety, we increased the value from 1.25 to 1.5, which reduces the acceptable Tsai-wu criterion to 0.66. Although we may consider margin of safety higher than 1.5, such high value is irrelevant in practice. Increasing the margin of safety yields the increment of failure region, so it results with the increased numbers of failures. In average of 10 replications, PhysCAL got 0.5 additional failures (six zero-failures), which is the least among considered methods, and achieved the predictive accuracy of 3.74 microinches. Random sampling got 0.6 additional failures, and achieved the predictive error of 5.16 microinches. Meanwhile, ALM and ALC achieved 8.3 and 7.9 failures, respectively. Likewise, SEGP was applicable only for the last three cases, and induced 3.3 failures from those cases.

\section{Summary}\label{se:conclusion}
For physics-constrained systems that are expensive-to-evaluate, failure-averse active learning is proposed in this paper. In order to achieve safe active learning under the regime of implicit physics constraints, GP priors are imposed on the target function and physics constraints, and two acquisition functions are developed for the safe variance reduction and the safe region expansion. For the safe variance reduction, two safe regions with different safety levels are employed in the IMSE criterion, thereby maximizing the safety and variance reduction over the underlying safe region. For the safe region expansion, the acquisition function is devised to sample near the safe region boundary considering uncertainty. Two acquisition functions are endowed with different objectives, so MOO framework with Pareto optimality is applied to integrate them into the flexible global criterion. The integrated acquisition function is sufficient for the Pareto optimality of the design point to be queried, and can be flexibly adjusted by decision maker's preference considering the trade-off between two acquisition functions. Also, it is shown that the integrated acquisition function asymptotically leads to the best estimation of system.

In the simulation study, the proposed approach showed promising performance with achievement of zero-failure, while the benchmark method fails to avoid failures in its learning process. Furthermore, with different parameters settings, we empirically observed that the safe variance reduction and the safe region expansion should be involved simultaneously for better predictive accuracy and higher safety. Our method has also shown remarkable performance in the case study, the predictive modeling of composite fuselage deformation considering its structural failure with Tsai-wu criterion. It achieved zero-failure in most cases, while other benchmark methods inducing more failures or inferior predictive accuracy. Our proposed method is adaptive, since it can incorporate domain knowledge and decision maker's preference with amenable parameters. Therefore, it is also applicable to other domains that are subject to implicit failures.



\appendices
\section{Proof of Theorem 1}\label{append:proof_theo1}
For any $\mf{x}_i\in S_i$, the probability of failure is
\begin{align*}
    &\text{Pr}\left(\hat{h}_i(\mf{x}_i) \ge \xi\,\big|\,\mf{x}_i\in S_{i-1}\right)\\
    &\le \text{Pr}\left(\hat{h}_i(\mf{x}_i) \ge \hat{\mu}^h_{i-1}(\mf{x}_i) + \beta_{i-1}\hat{\sigma}^h_{i-1}(\mf{x}_i)\,\big|\,\mf{x}_i\in S_{i-1}\right)\\
    &= p_{i-1},
\end{align*}
and it holds for $\forall\, i$. Then, by Boole's inequality (union bound), we have
\begin{align*}
    &\text{Pr}\left(\inf_i \left(\hat{h}(\mf{x}_i) \ge \xi\,\big|\,\mf{x}_i \in S_{i-1}\right)\right)\\
    &\le \sum_{i}\text{Pr}\left(\hat{h}_i(\mf{x}_i) \ge \xi\,\big|\,\mf{x}_i\in S_{i-1}\right)\\
    &\le \sum_{i} p_{i-1}.
\end{align*}
Suppose we use $p_i = p$ for $\forall\, i$ consistently. Then, the failure probability can be simplified as
\begin{align*}
    \sum_{i} p_{i-1} &= Np \triangleq \gamma\implies p = \frac{\gamma}{N}.
\end{align*}

\section{Proof of Theorem 2}\label{append:proof_theo2}
First, we show that $S_n$ converges to a nonempty subset of $\Omega_S$ an $n\to\infty$ by the consistency of GP regression over $\Omega$.
\begin{lemma}\label{lemma}
Let $S_{0}$ be the initial estimated safe region that contains a safe design point at least. With any dense sequence of design points in $S_n$ including PhysCAL, $S_n$ converges to a nonempty $S_\ast \subseteq \Omega_S$.
\begin{proof}
Let us denote a union of estimated safe regions as
\begin{align*}
    T_n = \bigcup_{i=0}^{n} S_i,
\end{align*}
which is bounded by $\Omega$. Suppose any dense sequence of design points $\{x_1,\,x_2,\ldots,\,x_n\} \subset T_n$. Since the sequence is monotone, and bounded by $\Omega$, it converges almost surely to $\Omega$ as $n\to\infty$. Note that the collection of dense sequences of design points includes PhysCAL, so the consistency of GP regression holds over $\Omega$ \cite{koepernik2021consistency}. Thus, $\hat{f}_n$ and $\hat{h}_n$ converge almost surely to the best MSE predictor $\hat{f}_\ast$ and $\hat{h}_\ast$ associated with $\Omega$. Since $\hat{h}_\ast$ is unbiased and has zero-variance over $\Omega$, for a large enough $n$, we have that
\begin{align*}
    S_n &= \{\mf{x}\in \Omega|\;\hat{\mu}^h_\ast(\mf{x}) < \xi\}\\
    &= \{\mf{x}\in \Omega|\;h(\mf{x}) < \xi\}\\
    &\triangleq S_\ast \subseteq \Omega_S,
\end{align*}
where $\hat{\mu}^h_\ast$ is the mean function of $\hat{h}_\ast$.
\end{proof}
\end{lemma}

Lemma \ref{lemma} allows us to utilize the consistency of GP models to show that degeneracy of $J(\mf{x})$. Recall that both $J_f$ and $J_h$ are nonnegative, and they are upper bounded by 
\begin{align*}
    J_f(\mf{x})&\le \mb{V}_{S_n^+}(\hat{f}_n) \le \sup_{\mf{x}\in \Omega}\,\text{Var}\left(\hat{f}_n(\mf{x})\right),\\
    J_h(\mf{x})&\le \eta^2(\mf{x}) \le \sup_{\mf{x}\in\Omega}\, \alpha^2 \text{Var}\left(\hat{h}_n(\mf{x})\right),
\end{align*}
for any $\mf{x}\in \Omega$. Then, by the consistency of GP, we have
\begin{align*}
    \lim_{n\to\infty}\sup_{\mf{x}\in\Omega}\text{Var}\left(\hat{f}_n(\mf{x})\right) &= 0,\\
    \lim_{n\to\infty}\sup_{\mf{x}\in\Omega}\text{Var}\left(\hat{h}_n(\mf{x})\right) &= 0.
\end{align*}
Thus, by the squeeze theorem, $J(\mf{x})$ converges to zero for every $\mf{x}\in \Omega$.

\bibliographystyle{IEEEtran}
\bibliography{arxiv_physcal.bib}

\end{document}